\PassOptionsToPackage{table}{xcolor}
\documentclass[numbered]{trbunofficial}
\usepackage[ruled,vlined]{algorithm2e}
\usepackage{algpseudocode}
\usepackage{amsmath}
\usepackage{amssymb}
\usepackage{graphicx} 
\usepackage{mathrsfs}
\usepackage{mathtools}
\usepackage{tikz}
\usepackage{float}
\usepackage[colorlinks=false,linkcolor=blue,citecolor=blue]{hyperref}
\usepackage{subcaption}
\usetikzlibrary{shapes,arrows}

\graphicspath{ {./Images/} }
\usepackage{subcaption}
\usepackage[section]{placeins}
\usepackage{booktabs, multirow} 
\usepackage{soul}
\usepackage{lipsum}
\usepackage{booktabs, multirow} 
\usepackage{soul}
\usepackage[table]{xcolor} 
\usepackage{changepage,threeparttable}
\usepackage{multicol}

\AuthorHeaders{Leong \& Abdelhalim et al.}
\title{MetRoBERTa: Leveraging Traditional Customer Relationship Management Data to Develop a Transit-Topic-Aware Language Model}

\author{\textbf{Michael Leong\textsuperscript{*}}\\
  Department of Urban Studies and Planning \\
  Massachusetts Institute of Technology, Cambridge, MA 02139, USA \\
  Email: mleong1@mit.edu \\
  \hfill\break
  \textbf{Awad Abdelhalim\textsuperscript{*}}, Corresponding Author\\
  Department of Urban Studies and Planning \\
  Massachusetts Institute of Technology, Cambridge, MA 02139, USA \\
  Email: awadt@mit.edu \\
  \hfill\break
  \textbf{Jude Ha} \\
  School of Engineering and Applied Sciences \\
  Harvard University, Cambridge, MA 02138, USA \\
  Email: jhchang@college.harvard.edu \\
  \hfill\break
  \textbf{Diane Patterson} \\
  Senior Performance Analyst, Washington Metropolitan Area Transit Authority \\
  300 7th Street SW, Washington, DC 20024, USA \\
  Email: DPatterson1@wmata.com \\
  \hfill\break
  \textbf{Gabriel L. Pincus} \\
  Data Scientist, Washington Metropolitan Area Transit Authority \\
  300 7th Street SW, Washington, DC 20024, USA \\
  Email: GLPincus@wmata.com\\
  \hfill\break
  \textbf{Anthony B. Harris} \\
  Senior Program Manager, Washington Metropolitan Area Transit Authority \\
  300 7th Street SW, Washington, DC 20024, USA \\
  Email: ABHarris@wmata.com\\
  \hfill\break
  \textbf{Michael Eichler} \\
  Strategic Planning Advisor, Washington Metropolitan Area Transit Authority \\
  300 7th Street SW, Washington, DC 20024, USA \\
  Email: meichler@wmata.com\\
  \hfill\break
  \textbf{Jinhua Zhao} \\
  Department of Urban Studies and Planning \\
  Massachusetts Institute of Technology, Cambridge, MA 02139, USA \\
  Email: jinhua@mit.edu 
}

\TotalWords{4993}
\nolinenumbers

\begin{document}
\maketitle
\section{Abstract}
Transit riders' feedback provided in ridership surveys, customer relationship management (CRM) channels, and in more recent times, through social media is key for transit agencies to better gauge the efficacy of their services and initiatives. Getting a holistic understanding of riders' experience through the feedback shared in those instruments is often challenging, mostly due to the open-ended, unstructured nature of text feedback. In this paper, we propose leveraging traditional transit CRM feedback to develop and deploy a transit-topic-aware large language model (LLM) capable of classifying open-ended text feedback to relevant transit-specific topics. First, we utilize semi-supervised learning to engineer a training dataset of 11 broad transit topics detected in a corpus of 6 years of customer feedback provided to the Washington Metropolitan Area Transit Authority (WMATA). We then use this dataset to train and thoroughly evaluate a language model based on the RoBERTa architecture. We compare our LLM, \emph{MetRoBERTa}, to classical machine learning approaches utilizing keyword-based and lexicon representations. Our model outperforms those methods across all evaluation metrics, providing an average topic classification accuracy of 90\%. Finally, we provide a value proposition of this work demonstrating how the language model, alongside additional text processing tools, can be applied to add structure to open-ended text sources of feedback like Twitter. The framework and results we present provide a pathway for an automated, generalizable approach for ingesting, visualizing, and reporting transit riders' feedback at scale, enabling agencies to better understand and improve customer experience.


\vspace{10pt}
\noindent\textit{Keywords}: Public Transport, Large Language Models, Customer Feedback.

\newpage
\section{Introduction}

\par Public transportation agencies collect a wealth of customer experience data through traditional customer relationship management (CRM) feedback channels, ridership surveys, and social media interactions. As transit agencies continue to place more emphasis on customer experience to recover post-pandemic ridership, collecting and timely analyzing feedback is of utmost importance for making data-informed decisions to improve customer satisfaction and attract more riders \cite{masstransit2022}. While the classical focus of customer service staff - in transit agencies and elsewhere, has been to review and respond to individual customer comments through traditional CRM channels, there is an ever-growing need to adapt to the use of social media, and especially Twitter, as a key channel of customer feedback \cite{marolt2015social}. The unstructured nature of this feedback, however, combined with the lack of transit-specific natural language models and tools, presents difficulties for agencies and public transit researchers hindering their ability to efficiently and meaningfully analyze feedback for transit-specific trends over time and at scale.

\par While conventional natural language processing (NLP) tools can be applied to open-text customer feedback in existing form, the broad range of topics in these comments and their generic aggregations does not lead to meaningful analysis. This is especially true for uncategorized data such as open-ended survey responses, call center transcripts, or tweets. Looking at the traditional use of NLP for sentiment analysis as an example, while the distribution of positive to negative comments could provide a proxy for overall customer experience, false equivalencies are drawn between the severity, urgency, or resolvability of comments. In the context of a transit agency, for example, conducting a topic-agnostic sentiment analysis to assess the impact of a presumably favorable new fare policy could be falsely impacted by an influx of negative sentiment feedback due to an incident-driven service disruption that occurred during the same period of analysis. To help address this gap, we conducted this study analyzing all instances of CRM feedback at the Washington Metro Area Transit Authority (WMATA) from 2017 to 2022 to create an end-to-end framework for meaningfully categorizing customer feedback and interactions, allowing for more robust transit service and policy evaluation. The language model we develop and present as a result of this work, \emph{MetRoBERTa}, is a large language model (LLM) which can provide a predictive classification of 11 transit-specific topics. This classification model, supplemented with existing NLP tools for text mining and sentiment analysis, allows for the meaningful aggregation of customer feedback about specific topics at scale. It also paves a path that could enable the automatic creation and routing of action items from customer feedback to transit agency departments, overcoming the time-intensive and subjective process of manually categorizing and routing rider feedback.



\subsection{Objectives and Contribution}

In this study, we propose and evaluate a framework for detecting latent transit-specific topics and classifying unstructured transit customer feedback into those topics. To the authors' best knowledge, this work is the first to efficiently leverage historic transit agency CRM data to develop an end-to-end, transit-topic-aware natural language model. We demonstrate the high accuracy of the resulting model in classifying unstructured customer feedback, using data from Twitter as a case study, and further re-classifying feedback from existing CRM channels that may have been misclassified during the traditional manual input process. The following sections of this paper are as follows: (a) a survey of related works, (b) a detailed breakdown of the CRM data preprocessing, modeling, and language model training (c) results, including model evaluation and potential case study applications using data from Twitter, and, (d) discussion and conclusions.


\section{Related Work}
Natural Language Processing is a field of artificial intelligence focusing on the interaction between computers and human language. It enables computers to process, analyze, and derive insights from vast amounts of textual data, including text classification, named entity recognition, information extraction, sentiment analysis, translation, prompt answering, and text summarization. Recent advances in NLP have made it possible and accessible for domains outside computing to utilize NLP to better understand the customer experience by mostly focusing on analyzing the \emph{sentiment} of customer feedback \cite{nguyen2015stocks, Rane_2018airlines, Philander_2016hospitality}. 

Naturally, transit agencies are posed to benefit from such advances. Understanding rider attitudes and the sentiment is paramount for transit agencies to assess and improve service, especially at a time when agencies are struggling to recover ridership to pre-pandemic levels \cite{watkins2021recent}. Social media, and Twitter (recently rebranded to "$X$") in particular, offer agencies the opportunity to engage with large amounts of customer feedback\footnote{The ongoing changes with the Twitter user interface and API may change this in the future \cite{aptwitter}.}, often in real or near-real-time, supplementing the traditional methods of infrequent and scope-confined ridership surveys \cite{Lock_2020}. While extracting meaningful customer experience metrics for a specific domain like public transit introduced the need for some pre-processing of the unstructured feedback data (e.g. creating baseline topics), the efficacy of this newfound data source has been demonstrated in numerous recent studies. Das and Zubaidi \cite{Das_2021} collect transit-related tweets from New York (NY) and California (CA) using a list of significant key terms (e.g. @NYTransit) and conducted sentiment and emotion analysis on the tweets using open-source libraries (i.e. no transit-specific lexicons developed). They concluded that CA and NY have different emotional content in their transit-related tweets, indicating that binary sentiment analysis may be inadequate in analyzing transit tweets where specific sentiment lexicons may be needed. Hosseini et al. \cite{Hosseini_2018} proposed linking semantic analysis with social network analysis to better understand transit customer satisfaction. In addition to collecting large amounts of Tweets, the authors identified accounts that are strictly transit-specific, accounts that are influential, and further classified accounts into the general public, reporters, unofficial transit bloggers, transit officials, and politicians. They also incorporated transit-specific lexicons using terms identified from the Transit Cooperative Research Program. The study demonstrated that customers are more likely to utilize Twitter for real-time concerns (e.g. safety, travel time, service, and information provision) compared to planning topics (e.g. spatial and temporal availability) and that topic discussions vary depending on service status and locale. Kabbani et al. \cite{Kabbani_2022} conducted a similar study which reciprocated the results, concluding that key incidents can be identified from Twitter feedback, allowing agencies to quickly identify and respond to issues in near real-time.

More recently, researchers focused on the means of utilizing Twitter feedback to assess and improve operations. Haghighi et al. \cite{Haghighi_2018} and Al-Sahar et al. \cite{al2023using} conducted studies concluding that utilizing customer-oriented measures from Twitter to supplement operational-oriented measures from traditional transit data sources would empower agencies to improve planning and operational decisions \cite{Liu_2019}. While findings from studies conducted in different cities show consistent results on the value-added of language processing for extracting useful information on transit riders' experience, they also identify limitations associated with traditional NLP methods like the need for standardization and lemmatization \cite{Mendez_2019chile, Luong_2015la}, the difficulty of incorporating geocoding and topic modeling for the personal, decontextualized, and specific nature of tweet texts \cite{Arjona_2020madrid}, and that word-frequency-based methods are limited by differing local contexts \cite{Das_2021}.

The advent of language models in recent years has provided immense contributions to advancing state-of-the-art natural language processing. Large Language Models (LLMs), most notably BERT \cite{devlin2018bert}, GPT \cite{radford2019gpt}, and their derivatives, have proven extremely useful for general-purpose text mining and analysis tasks. In this study, we propose utilizing existing customer feedback from traditional sources to develop a LLM capable of inferring transit-specific topics. We additionally assess the value-added of this approach in identifying and classifying the transit-specific topic of tweets in a highly generalizable, end-to-end manner.


\section{Methodology}

\begin{figure}[!h]
    \centering
    \includegraphics[width=0.99\textwidth]{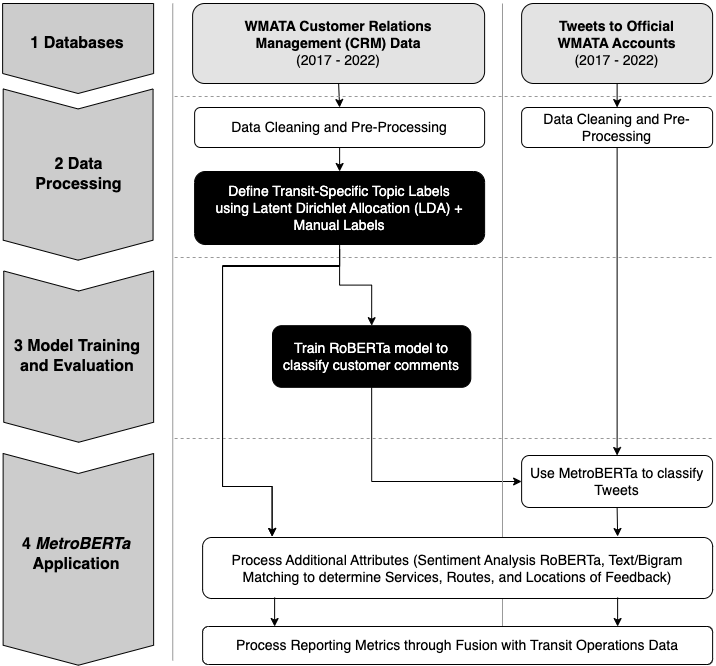}
    \caption{Data sources, processing, training, and application steps for the proposed framework.}
    \label{fig:flowchart}
\end{figure}

\newpage
The data sources, steps for preprocessing, model training, and application for analysis are illustrated in Figure \ref{fig:flowchart}. The main data source used in this study is a dataset containing a total of $\sim$180,000 customer feedback comments with manually-classified labels from WMATA's CRM database. This dataset includes customer complaints and feedback from January 2017 through December 2022. The dataset has been anonymized for any Personal Identifiable Information (PII) by WMATA before being made accessible to the researchers. The curated dataset includes the text written by the customers or customer service agent in the case of comments made over the phone or online chat, in addition to a manually selected label that maps the complaint to one of 61 \emph{problem categories}. The developed language model was applied to a second dataset containing $\sim$300,000 tweets corresponding to the same period of CRM data. Those were only direct tweets that mentioned one of WMATA's official Twitter handles (namely @wmata, @wmatagm, @metrobusinfo, and @metrorailinfo).

\subsection{Detecting Transit-Specific Topics from Customer Feedback}

Although WMATA's CRM data contained human-classified category labels which were used to route action items from feedback to internal departments, the large number of distinct labels that every feedback instance can be assigned to (61 categories) makes it challenging to directly utilize the dataset as-is for training a classification model. While it could be possible to manually create a hierarchical, parent-child structure to condense the number of categories, there is no guarantee that the customers would consistently assign similar issues to fall within the same \emph{problem category}. As such, we opted to utilize a semi-supervised learning algorithm, Latent Dirichlet Allocation (LDA), on the dataset to re-categorize customer feedback instances into fewer, broader latent topics based on the co-occurrence of specific lexicon within the comments' text. WMATA's CRM \emph{problem categories} are fully listed in Appendix I.

\begin{figure}[!h]
    \centering
    \includegraphics[width=0.95\textwidth]{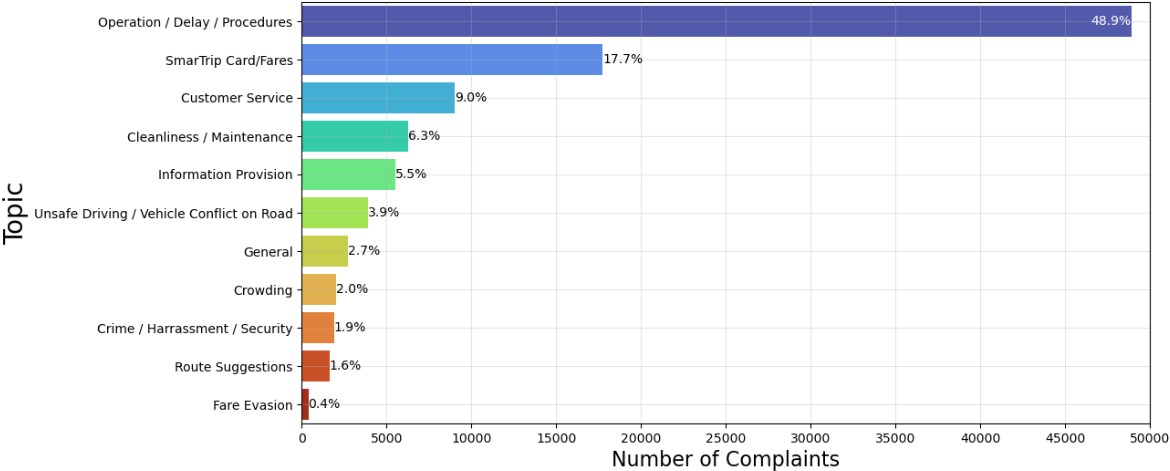}
    \caption{Distribution of topics across CRM complaints in the training dataset.}
    \label{fig:training_topics}
\end{figure}

Prior to applying LDA to the dataset, hierarchical clustering was employed as a screening mechanism to identify an upper limit of potential topics, which was found to be 23 (i.e. 23 broad clusters best represented the 61 manually-labeled individual problem categories). We then ran a 23-topic LDA, which yielded broad transit-specific topics, the words associated with those topics, and their relative contribution to the topic score. The authors identified topics with similar keywords which were further condensed to result in 9 topics (e.g. individual topics of delays, not stopping, and early departures were condensed into a single topic of delays and operations). A limitation of LDA, however, was being unable to detect niche topics that were significantly less represented compared to others. This includes critical topics such as complaints regarding crime, harassment, and safety in the transit system. Additionally, $\sim$62,000 (about 35\%) complaints did not have a significant primary topic score and were left unassigned, as the keywords detected in their text could map to multiple latent topics. To resolve the former issue, the niche topics of crime, harassment, and security were held out of the CRM dataset prior to applying LDA and categorized as a single topic based on the original manually-assigned CRM labels. The same was done for the crowding complaints as a topic of its own. The resulting set of $\sim$120,000 complaints that had a strong primary topic score (i.e. keywords detected in the complaint text map strongly towards one topic more than others) was assigned to 11 topics (9 LDA-detected topics and the 2 manually categorized topics). The complaint text and LDA topic label were used as a training dataset, illustrated in Figure \ref{fig:training_topics}. Table \ref{tab:topics} shows those topics, their keywords, and relevant categories from the 61 \emph{problem categories} manually assigned at complaint entry. Appendix II shows the ratio that CRM problem categories contribute to each LDA topic (up to the top 3 categories per topic, for brevity).

\begin{algorithm}[!h]
  \SetAlgoLined
  \DontPrintSemicolon
  \KwIn{WMATA CRM dataset, number of seed topics $K=23$}
  \KwOut{9 condensed topics using Latent Dirichlet Allocation (LDA)}
  
  \textbf{Step 1: Data Preprocessing} \\
  \Indp
  Tokenize the complaints into words\;
  Remove stop words, punctuation, and WMATA-specific terms\;
  Perform stemming \& lemmatization\;
  Create a complaint-word matrix\;
  \Indm

  \textbf{Step 2: Model Training} \\
  \Indp
  Initialize LDA model with $K$ topics\;
  Randomly assign topics to words in the complaint-word matrix\;
  \Indm
  \While{not converged}{
    \ForEach{complaint $d$}{
      \ForEach{word $w$ in $d$}{
        Calculate probability of word $w$ belonging to each topic $t$\;
        Assign word $w$ to a topic based on the probability\;
      }
    }
    Update topic-word and complaint-topic distributions\;
  }
  \textbf{Step 3: Topic Inference} \\
  \ForEach{complaint $d$}{
    Infer the topic distribution for complaint $d$\;
  }
  \textbf{Step 4: Topic Interpretation} \\
  \ForEach{topic $t$}{
    Identify top words in topic $t$\;
    Interpret topic $t$ based on the top words\;
  }

  \caption{LDA for Re-Categorizing Customer Feedback Complaint Categories}
  \label{alg:tf_idf}
\end{algorithm}


\clearpage
{\renewcommand{\arraystretch}{1.5}
\begin{adjustwidth}{-2.5 cm}{-2.5 cm}\centering\begin{threeparttable}[t]\centering
\caption{Broad Transit-Specific Topics and Keywords}\label{tab:topics}
\scriptsize
\begin{tabular}{|p{0.7in}|p{0.45in}|p{3in}|p{1.7in}|}
\hline
\textbf{Broad Topic} &\textbf{Detection Method} &\textbf{Keywords} &\textbf{Relevant WMATA CRM Categories} \\\hline
Operations, Delays,\newline Procedures &LDA &minutes late, late work, never showed, station manager, says operator, next stop, minutes early, people waiting, never came, failed stop, didn't stop, two buses, kept going, scheduled time &Delays, Did Not Stop, No Show, Following Procedures, Early, Service Disruption, Rush Hour Promise, Service Changes, Did Not Follow Route, Door Closing \\
SmarTrip Card, Fares &LDA &smartrip card, credit card, card number, new card, stored value, money card, add value, charged twice, monthly pass, day pass, vending machine, money back, apple wallet &SmarTrip Card, Fare Value Missing, Incorrect Fare, Incorrect Parking Fee, Faregate, Farebox, Fare Increase \\
Customer \newline Service &LDA &station manager, dont know, first time, make sure, commend driver, even though, asked driver, extremely rude, job well, excellent service, great job, went beyond, another passenger, driver said &Commendation, Rude Employee, Unprofessional Behavior, Passenger Assistance, Unprofessional Tone \\
Cleanliness, Maintenance &LDA &air conditioning, rail cars, smells like, lower level, escalator going, street level, looks like, needs cleaning, degrees outside, safety issue, escalator working, station entrance &Maintenance, Cleaning, Cooling, Lighting, Heat, Restrooms \\
Information Provision &LDA &rush hour, line trains, wait minutes, track work, single tracking, every day &Inaccurate Information, Bus Info Display, Announcements, Platform Info Display, Train Info Display \\
Unsafe \newline Driving &LDA &right lane, left lane, left turn, almost hit, oncoming traffic, right turn, speed limit, ran light, bike lane, slam brakes, driver cut, hit car &Unsafe Operation, Curbing Bus \\
General &LDA &trip planner, vintage metrobuses, metropolitan area, new look, reserved parking, parking lot, flexible new, good morning &General Information, Other, Suggestion, Trip Planner, Injury, Advertising, Property Damage, Bicycles \\
Route \newline Suggestions &LDA &metrobus routes, new routes, new metrobus, need new, add new, need reroute, reroute station, add express &Request Additional Service \\
Fare Evasion &LDA &fare evasion, without paying, paying fare, fare gates, people jumping, every time, every day, fare evaders, jumping turnstiles, don't pay &Fare Evasion \\
Crowding &Manual &rush hour, realtime crowding, many people, standing room, social distancing, morning rush, every day, during peak, six car, people waiting &Crowding \\
Crime, \newline Harassment, Security &Manual &transit police, parking lot, police officer, young man, feel safe, police presence, people smoking, homeless people, second time &Crime, Harassment, Security \\
& & & \\\hline
\multicolumn{4}{|c|}{WMATA Categories which span Multiple Topics: Safety, MetroAccess, Crowding, Mobile Feedback, Accessibility} \\
\hline
\end{tabular}
\end{threeparttable}\end{adjustwidth}
}

\subsection{Topic Classification Models}
\subsubsection{Traditional Machine Learning Models}

To develop a topic classification model that utilizes the LDA-labelled dataset, we use Term Frequency Inverse Document Frequency (TF-IDF) to create a matrix representation of the customer complaints and the terms used across different topics. The term frequency (TF) provides a measure of the importance of a term (t) within a specific customer complaint (d). While the inverse document frequency (IDF) is a measure of the uniqueness of term (t) across a collection across the entire collection of complaints (D). By multiplying the TF and IDF, higher weights are assigned to terms that are frequent within a specific topic but relatively rare across all complaints.

\begin{equation}
\text{TF}(t, d) = \frac{\text{Number of occurrences of term } t \text{ in complaint } d}{\text{Total number of terms in complaint } d}
\label{eq:tf}
\end{equation}

\begin{equation}
\text{IDF}(t, D) = \log\left(\frac{\text{Total number of complaints in the CRM dataset } D}{\text{Number of complaints containing term } t}\right)
\label{eq:idf}
\end{equation}

\vspace{10pt}
\noindent Combining TF and IDF:

\begin{equation}
\text{TF-IDF}(t, d, D) = \text{TF}(t, d) \times \text{IDF}(t, D)
\label{eq:tf-idf}
\end{equation}

\vspace{10pt}
\noindent To create a TF-IDF feature matrix, we calculated the frequency of unigrams (single-word terms) and bigrams (two-word terms) that have appeared at least 20 times across all CRM complaints. This was done after accounting for stop words and WMATA-specific terms (route and station names etc.). The resulting TF-IDF feature matrix that was used as an input to the classification models contained weights of approximately 20,000 unique terms.

Using the TF-IDF feature matrix as an input, we trained and evaluated 5 traditional machine learning (ML) models: a Random Forest classifier, a Linear Regression with Stochastic Gradient Decent (SGD), a Support Vector Machine, Naive Bayes, and a Logistic Regression classifier. All models were trained on 5-fold cross-validation with a training-validation split of 80-20. In calculating the loss function (multi-class cross-entropy), we used an inverse-weight vector to account for the differences in topic sizes illustrated in Figure \ref{fig:training_topics}, assigning a higher weight to the lower-represented classes to prevent the model from biasing towards over-represented topics.

\subsubsection{Masked Language Modeling}
While TF-IDF offers a powerful way to represent a text corpus for training and deploying traditional machine learning algorithms, a key limitation is that it struggles with out-of-vocabulary (i.e. previously unseen) terms. While the term feature matrix extracted from our dataset contains $\sim$20,000 unique terms, any terms not seen in this glossary would constitute a null value to the underlying predictive model. Another key limitation is that TF-IDF-based models do not take context into account, and rather treat individual terms (unigrams and bigrams in this study) independently. 

Given that one of the core areas of interest for WMATA is creating a model that supports extracting and aggregating feedback from unstructured data sources like Twitter, having a model capable of semantic and contextual understanding is key. Additionally, once adequately trained, LLMs' ability to transfer-learn allows for the creation of a generalizable transit-topic-aware model that is not restricted by the spatial and temporal use of certain terms to describe transit topics. The core LLM we utilized for this study is the RoBERTa model open-sourced by Facebook (now Meta) Research \cite{liu2019roberta}, re-trained to our dataset as described in Algorithm \ref{alg:roberta}. We additionally compared a model variant that was further pre-trained on $\sim$124M tweets and fine-tuned for multi-label topic classification \cite{antypas2022twitter}, but found the RoBERTa base model generalized better to our data.

\begin{algorithm}[!h]
  \SetAlgoLined
  \DontPrintSemicolon
  \KwIn{LDA-labelled CRM Dataset}
  \KwOut{RoBERTa-based transit topic classification model}
  \KwData{80-20 train-test split, 11 LDA-processed labels}

  \textbf{Step 1: Data Preprocessing} \\
  \Indp
  Tokenize the complaints into words\;
  Remove stop words and punctuation\;
  Remove WMATA-specific terms (route and station names etc.)\;
  Perform stemming \& lemmatization\;
  Encode the complaints using RoBERTa's tokenizer\;
  \Indm

  \textbf{Step 2: Train-test Split} \\
  \Indp
  Split the dataset into training and testing sets with an 80-20 train-test ratio\;
  \Indm

  \textbf{Step 3: Model Initialization} \\
  \Indp
  Initialize the RoBERTa with pre-trained weights\;
  Add a classification layer on top of the LLM\;
  Randomly initialize the classification layer weights\;
  \Indm

  \textbf{Step 4: Model Training and Evaluation} \\
  \Indp
  Log and observe training loss curves\;
  Fine-tune the RoBERTa hyperparameters to improve training\;
  Evaluate the trained model on the testing set\;
  Calculate relevant performance metrics (accuracy, precision, recall, and F-1 score) for the 11 distinct topics\;
  Create and save new model checkpoints for future deployment\;
  \Indm

  \caption{Training a Topic Classification Model using RoBERTa LLM}
  \label{alg:roberta}
\end{algorithm}

\subsection{Extended Analysis Pipeline}
The topic classification model was designed to fit within a pipeline of broader NLP tools, fused with transit operations data in order to aggregate performance metrics. After assigning a transit topic to a feedback instance, we additionally (to the extent possible) extract the following:

\begin{itemize}
  \item \emph{Sentiment} using open-source RoBERTa sentiment model \cite{barbieri-etal-2020-tweeteval}.
  \item \emph{Mode, route, vehicle, and location} using uni/bigram matching to WMATA assets.
  \item \emph{Customer characteristic} (e.g. gender or frequency of feedback) using text inference.
  \item \emph{Normalization} of feedback volume based on daily ridership data.
\end{itemize}

\section{Results and Discussion}

\subsection{Language Model Evaluation}
\begin{figure}[!h]
    \centering
    \includegraphics[width=0.90\textwidth]{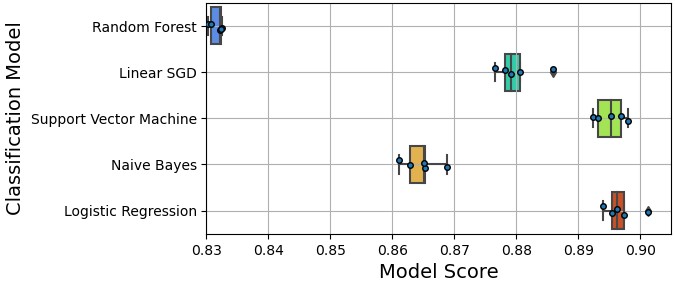}
    \caption{Performance of TF-IDF-based ML classification models.}
    \label{fig:tf-idf_classifiers}
\end{figure}

Figure \ref{fig:tf-idf_classifiers} shows the accuracy of the traditional ML models evaluated, with the box plots illustrating the score distribution and the scatters showing the individual score of each of the 5-fold training.

\begin{figure}[!h]
    \centering
    \includegraphics[width=0.99\textwidth]{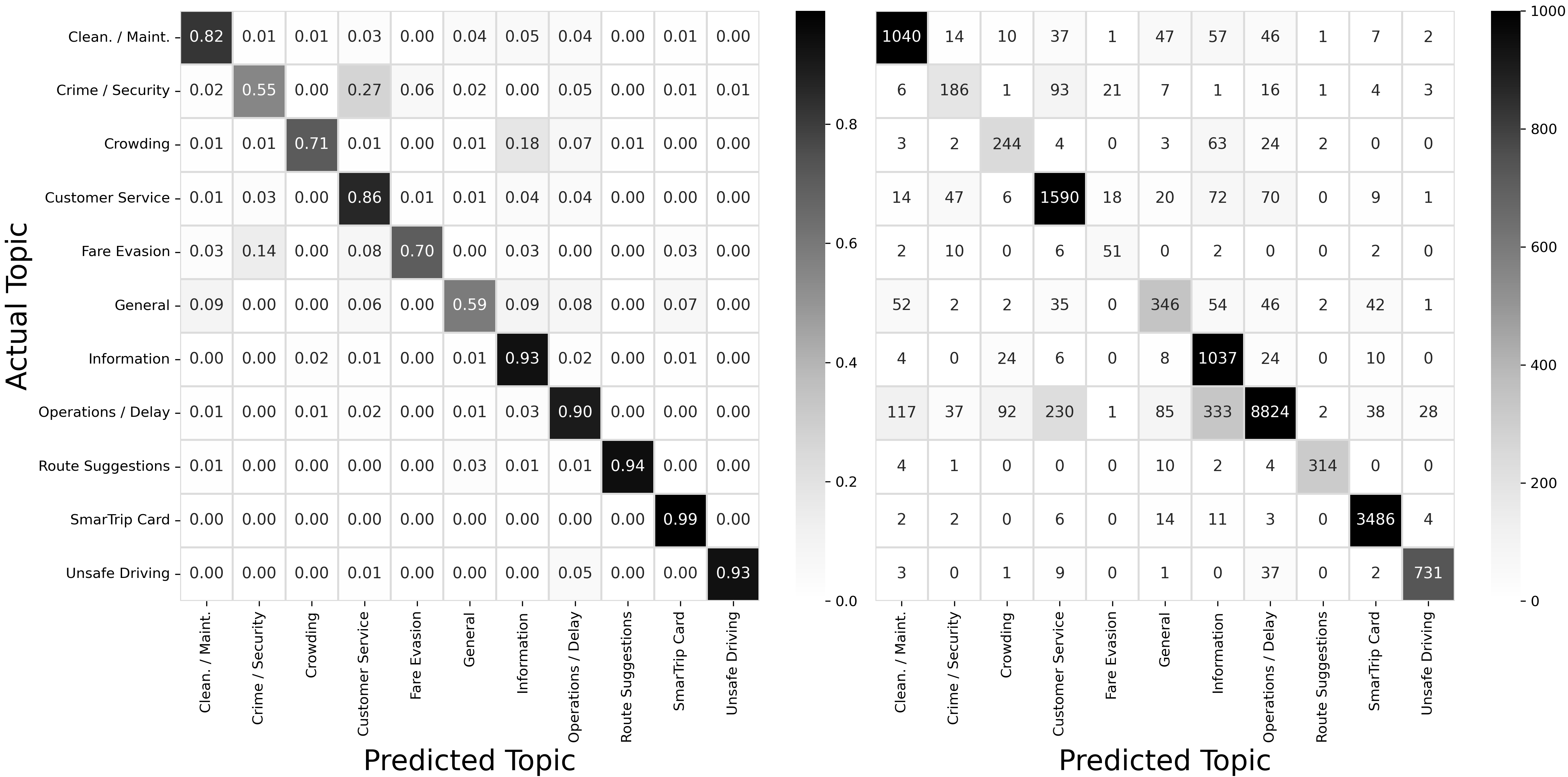}
    \caption{\emph{MetRoBERTa} topic classification accuracy (left) and raw counts (right).}
    \label{fig:roberta_classification}
\end{figure}

The Logistic Regression model was found to be the best performer. We compare its performance to \emph{MetRoBERTA} which was trained on the same dataset as elaborated in Algorithm \ref{alg:roberta}. Table \ref{tab:model_comparison} shows the comparison between the two models. The macro and weighted averages are respectively the mean of the raw metric scores for each topic and the topic-size weighted mean. Figure \ref{fig:roberta_classification} shows the confusion matrices for \emph{MetRoBERTa} in percentage points and raw inter-topic counts for the predicted topic classifications of the 20,000 feedback instances in the test subset.

\begin{table}[!h]
  \renewcommand{\arraystretch}{1.00}
  \centering
  \caption{Evaluation Metric Comparison for Best-Performing TF-IDF Model and MetRoBERTA}
    \begin{tabular}{|l|c|c|c|c|c|c|}
    \hline
    \textbf{Classification Model $\rightarrow$} & \multicolumn{3}{c|}{\textbf{TF-IDF Logistic Regression}} & \multicolumn{3}{c|}{\textbf{MetRoBERTa}} \\
    \hline
    \textbf{Metric$\rightarrow$} & \multirow{2}[4]{*}{\textit{\textbf{Precision}}} & \multirow{2}[4]{*}{\textit{\textbf{Recall}}} & \multirow{2}[4]{*}{\textit{\textbf{F1-score}}} & \multirow{2}[4]{*}{\textit{\textbf{Precision}}} & \multirow{2}[4]{*}{\textit{\textbf{Recall}}} & \multirow{2}[4]{*}{\textit{\textbf{F1-score}}} \\
\cline{1-1}    \textbf{Topic $\downarrow$} &       &       &       &       &       &  \\
    \hline
    Cleanliness / Maintenance & 0.88  & 0.84  & 0.81  & 0.87  & 0.81  & 0.84 \\
    \hline
    Crime / Harassment / Security & 0.55  & 0.49  & 0.52  & 0.69  & 0.56  & 0.62 \\
    \hline
    Crowding & 0.52  & 0.73  & 0.61  & 0.79  & 0.71  & 0.72 \\
    \hline
    Customer Service & 0.81  & 0.80  & 0.77  & 0.78  & 0.86  & 0.81 \\
    \hline
    Fare Evasion & 0.45  & 0.80  & 0.58  & 0.69  & 0.70  & 0.62 \\
    \hline
    General & 0.78  & 0.37  & 0.50  & 0.64  & 0.59  & 0.62 \\
    \hline
    Information Provision & 0.73  & 0.85  & 0.78  & 0.68  & 0.93  & 0.76 \\
    \hline
    Operation / Delay / Procedures & 0.93  & 0.93  & 0.93  & 0.97  & 0.90  & 0.93 \\
    \hline
    Route Suggestions & 0.95  & 0.94  & 0.95  & 0.98  & 0.94  & 0.97 \\
    \hline
    SmarTrip Card / Fares & 0.93  & 0.99  & 0.97  & 0.93  & 0.99  & 0.98 \\
    \hline
    Unsafe Driving & 0.90  & 0.92  & 0.91  & 0.95  & 0.93  & 0.94 \\
    \hline
    \multicolumn{7}{|c|}{\hspace{2.25in}\textbf{Aggregate Evaluation Metrics}} \\
    \hline
    \textbf{Accuracy} & \multicolumn{3}{c|}{0.88} & \multicolumn{3}{c|}{\textbf{0.90}} \\
    \hline
    \textbf{Macro avg} & 0.77  & 0.77  & 0.76  & \textbf{0.81} & \textbf{0.81} & \textbf{0.80} \\
    \hline
    \textbf{Weighted avg} & 0.88  & 0.88  & 0.88  & \textbf{0.90} & \textbf{0.89} & \textbf{0.90} \\
    \hline
    \end{tabular}%
  \label{tab:model_comparison}%
\end{table}%

\emph{MetRoBERTa} outperforms the best-performing TF-IDF model across all aggregate evaluation metrics. Additionally, the semantic and context-aware nature of our BERT-based model allows it to correctly identify what were misclassifications either due to the manual nature of assigning problem categories by customers or customer service representatives, the keyword-based re-categorization of topics to create our training dataset through LDA, or both. Table \ref{tab:misclassification} shows examples where \emph{MetRoBERTa}'s prediction of the label did not match the ground truth label (LDA topic), and it provided a classification that would better fit the complaint. The \emph{problem category} is the manually-assigned label by the customer when submitting a CRM complaint. It can be seen in those examples that the TF-IDF model matches the LDA-assigned topic, which is reasonable given that both methods rely on keywords to detect topics. However, when words are used in a different context (e.g. pouring and swiping, which the keyword-based methods associated more with cleaning as in the first example) or when a customer inquires about the Smartcard and not "SmarTrip" card, keyword-based methods struggle. In cases like those demonstrated by the 2\textsuperscript{nd} and 5\textsuperscript{th} examples, the customers used words that are associated with operations and delays (e.g. buses don't show up, driver stopping) but the context of the comment was respectively better fitting to topics of information provision and customer service. Given that such instances negatively impact accuracy, the \emph{MetRoBERTa}'s true accuracy would be well above the reported 90\%. 

\begin{table}[!h]
\renewcommand{\arraystretch}{2.0}
\centering
\caption{Examples of \emph{MetRoBERTa} Out-Performing Lexicon-Based Methods}
\label{tab:misclassification}
\footnotesize
\begin{tabular}{|m{5cm}|m{3.0cm}|m{2cm}|m{1.9cm}|m{2.5cm}|}
\hline
\textbf{Sample Complaint} & \textbf{Problem Category} & \textbf{LDA Topic} & \textbf{TF-IDF} & \textbf{MetRoBERTa} \\
\hline
Bus completely cut me off in pouring rain. Came within an inch of side-swiping my car along the entire side passenger side. & Unsafe Operation & Cleanliness / Maintenance & Cleanliness / Maintenance & Unsafe Driving / Vehicle Conflict \\
\hline
On the 96 route I have seen a frustrating increase recently in the number of buses that don't show up on the One Bus Away system. & Inaccurate Information & Operation / Delay / Procedures & Operation / Delay / Procedures & Information Provision \\
\hline
There was supposed to be a bus at 7:37 and another 7:45. No bus came until 7:48. The bus was extremely crowded because we had all been waiting. & No Show & Operation / Delay / Procedures & Operation / Delay / Procedures & Crowding \\
\hline
How does a Senior obtain a Smartcard that provides for the Senior/Disability discount? & SmarTrip Card & General & General & SmarTrip Card / Fares \\
\hline
I was a half block away from the bus stop. That didn't stop the driver from stopping and letting me aboard. I appreciate her for doing that and WMATA for hiring her! & Commendation & Operation / Delay / Procedures & Operation / Delay / Procedures & Customer Service \\
\hline
\end{tabular}
\end{table}

\clearpage
\subsection{Value Proposition for Transit Agencies}
\subsubsection{Adding Structure to Unstructured Data}
\par Our proposed analysis pipeline featuring \emph{MetRoBERTa} combines the results of our transit-specific topic classification LLM with other attributes, such as sentiment, service, location, assets, and user characteristics, that can be obtained using text-mining and open-source tools. We provide a case study analyzing data from Twitter. After using our model to detect a topic, sentiment is inferred via a RoBERTa sentiment model \cite{barbieri-etal-2020-tweeteval}. Using a RoBERTa-based sentiment model is key for efficiency as both the sentiment and our classification model use the same text tokenizer. We used text mining to identify the mode of service, and match routes, stations, and vehicle IDs to a database of WMATA assets. A likely gender was inferred from the user's first name (when possible) using a method previously developed at WMATA \cite{shuman2022inferring}. Table \ref{tab: tweets} shows an application of this pipeline to add structure to tweets mentioning @wmata (and related WMATA accounts).

{\renewcommand{\arraystretch}{1.0}
\begin{adjustwidth}{-2.5 cm}{-2.5 cm}
\centering
\begin{threeparttable}[!htb]\centering
\caption{Sample Tweets Classification Using MetRoBERTa and Extended Analysis Pipeline}\label{tab: tweets}
\scriptsize
\begin{tabular}{|p{2.8 in}|p{0.7in}|p{0.5in}|p{0.35in}|p{0.85in}|p{0.35in}|}\hline
\textbf{Tweet} &\textbf{Detected Topic (MetRoBERTa)} &\textbf{Sentiment} &\textbf{Service} &\textbf{Route/Station/Car} &\textbf{Gender} \\\hline
@Metrorailinfo What's happening with the green line between West Hyattsville and PG Plaza for the last twenty-odd minutes? &Operation, \newline Delay, \newline Procedures &  Neutral &Rail &Green - West Hyattsville, Prince George's Plaza &Male \\\hline
@wmata train car 7507 needs to be cleaned. It has some makeup or something on the inside. It's in the glass and seat. Ruined a nice woman's clothes. &Cleanliness, Maintenance &Negative &Rail &Car 7507 &Female \\\hline
@Metrorailinfo Operator of Yellow line train to Huntington (Archives station at 8:49 am, containing car \#3200) was a delight this morning. Welcoming aboard and announcing each station clearly and with infectious joy. &Customer\newline  Service &Positive &Rail &Yellow - Archives &Male \\\hline
@WeMoveDC @wmata I live in Fairlawn and it's been so nice to be able to take a 30 bus from Pennsylvania Ave to 4th and Independence SW and it only take 10 minutes. There should be dedicated bus lanes there & Operation, \newline Delay, \newline Procedures & Positive & Bus& Route 30 - 4th \& Independence SW & Male \\\hline 
Hi @wmata the driver of the 70 Metrobus \#6032 needs a refresher on safe driving. He cut me and others off multiple times, speeding recklessly, while traveling north up 7th St NW at around 4:45 PM. &Unsafe \newline Driving, Road Conflict &Negative &Bus &Route 70, Vehicle 6032 &Male \\\hline
Even @wmata is getting into the spirit with commemorative fare cards!  That's how you KNOW it's a big day in DC.  Years from now, these cards might even be worth more than the \$ you load onto them.  Everything today is \#ALLCAPS   & SmarTrip Cards, Fares & Positive & - & - & Male \\\hline
@wmata @timkaine @MarkWarner Is anything going to be done about the gate jumpers who refuse to pay metro fares? I see them almost every time I take the metro. &Fare Evasion &Negative &Rail &- &- \\\hline
I'm at Gallery Place - Chinatown Metro Station - @wmata in Washington, DC https://t.co/rJy63TbK4B &General &Neutral &Rail &Red/Green/Yellow - Gallery Place &Male \\\hline
Car 5016 orange no air @Metrorailinfo gonna be a \#hotcar later today @MetroHotCars &Cleanliness, Maintenance &Neutral &Rail &Orange - Car 5016 &Female \\\hline
@wmata on train 3121 from New Carrollton to East Falls Church and someone is smoking weed &Crime, \newline Harassment, Security &Negative &Rail &Orange - East Falls Church, Car 3121 &- \\\hline
@wmata rush hour + 6 car trains + tourists = extremely dangerous situations on platforms &Crowding &Negative &Rail &- &- \\\hline
@kierig @wmata @Metrobusinfo @transitapp @Citymapper The @DCMetroandBus app. I highly doubt this is an issue with the app itself; more likely erroneous information from the bus tracking system. &Information Provision &Negative &Bus &- &- \\\hline
Hope to see more transit agencies connect \#micromobility services with transit offerings. @DavidZipper -- love this deep dive. FWIW @wmata's Bus Transformation Project makes this recommendation (s/o to the innovative thinking of @travelingali ) & General & Positive & - & - & Male \\\hline
\end{tabular}
\end{threeparttable}\end{adjustwidth}
}

\par The addition of a transit-specific topic as a dimension to the data allows for more meaningful aggregations of customer feedback across a multitude of transit agencies' interests. This will be briefly demonstrated in the following two applications: understanding sources of customer feedback and generating performance metrics from customer feedback.

\bigskip 

\subsubsection{Understanding Sources of Customer Feedback}

\par Tables \ref{tab: sent-topic} and \ref{tab: topic-mode} compare aggregations of WMATA's CRM and Twitter feedback (both from January 2017 to December 2022) by topic, sentiment, and mode. From the total numbers, we can observe that overall, there is more negative sentiment expressed in traditional CRM feedback compared to Twitter, where a higher percentage of neutral sentiment can be observed. This isn't surprising, considering that the majority of CRM feedback is obtained through what is officially called the "Customer \emph{Complaint} Process". Conversely, more neutral and slightly higher positive feedback is observed on Twitter. This may be attributable to the "comments in passing" nature of tweets, where the barrier to provide commentary on the go and tag relevant WMATA accounts is significantly lower than that of initiating an official customer complaint. 

\begin{adjustwidth}{-2 cm}{-2 cm}\centering\begin{threeparttable}[!htb]
\caption{CRM Feedback and Twitter Aggregated by Topic and Sentiment}\label{tab: sent-topic}
\renewcommand{\arraystretch}{1.30}
\small
\begin{tabular}{|p{1.7in}|p{0.5in} p{0.5in} p{0.5in}||p{0.5in} p{0.5in} p{0.5in}|}\hline
\multirow{2}{*}{\textbf{Topic}} &\multicolumn{3}{c}{\textbf{CRM Feedback}} &\multicolumn{3}{c|}{\textbf{Twitter}} \\\cline{2-7}
&Negative &Neutral &Positive &Negative &Neutral &Positive \\\hline
Cleanliness / Maintenance &73\% &23\% &4\% &45\% &42\% &13\% \\
Crime / Harassment &87\% &12\% &1\% &62\% &35\% &3\% \\
Crowding &71\% &17\% &12\% &52\% &38\% &11\% \\
Customer Service &56\% &10\% &34\% &33\% &51\% &16\% \\
Fare Evasion &83\% &15\% &2\% &60\% &36\% &4\% \\
General &43\% &48\% &9\% &21\% &64\% &15\% \\
Information Provision &81\% &14\% &5\% &55\% &34\% &11\% \\
Route Suggestions &22\% &74\% &4\% &8\% &87\% &6\% \\
Operations &76\% &22\% &2\% &44\% &48\% &8\% \\
SmarTrip Cards / Fares &70\% &27\% &3\% &41\% &46\% &14\% \\
Unsafe Driving &82\% &17\% &1\% &5\% &95\% &0\% \\
\hline
\textbf{Total (All Topics)} &75\% &17\% &8\% &51\% &38\% &11\% \\
\hline
\end{tabular}
\end{threeparttable}\end{adjustwidth}

\bigskip
The addition of topic classifications from MetRoBERTa, however, helps to further isolate the polarity of sentiments amongst different categories, as well as investigate how they vary between these two sources. For example, we are able to observe that the majority of positive feedback in the CRM data comes from customer service, specifically from a sub-category labeled "commendation", and the rate of positive customer service feedback on the CRM database actually exceeds that of Twitter.  The topics of Crime/Harassment/Security and Fare Evasion are seen to be consistently negative in sentiment in both CRM and Twitter, while topics of Cleanliness/Maintenance, Crowding, Information Provision, Operations, Fares, and Unsafe Driving tended to be more neutral and less negative on Twitter compared to CRM.

Looking at the modal breakdown in Table \ref{tab: topic-mode}, we observe that a significantly higher proportion of CRM complaints are bus-related compared to Twitter, which skews more towards rail service and general discussions on current affairs. Adding topic labels from MetRoBERTa, however, we are able to further segment that while $\sim$50\% of CRM feedback is regarding operations, we observe a better distribution of topics in the Twitter feedback, with customer service being similarly represented to operations, and other topics more equitably distributed. This underscores a large difference in the nature of feedback and demographics of feedback providers between the two sources. Additionally, we are able to isolate the bias in feedback frequency towards bus users on CRM and rail users on Twitter to the topics of Crowding, Operations, and Unsafe Driving, while the topics of Cleanliness/Maintenance, Crime/Harassment/Security, Information Provision, Route Suggestions, and SmarTrip Cards/Fares actually demonstrate a similar distribution of rail-to-bus comments despite the overall biases. These insights demonstrate how our topic classification model can be used to add an additional dimension to customer feedback analysis and draw actionable insights.

\bigskip
\begin{adjustwidth}{-2.5 cm}{-2.5 cm}\centering\begin{threeparttable}[!htb]\centering
\caption{CRM and Twitter Feedback Aggregated by Topic and Mode}\label{tab: topic-mode}
\renewcommand{\arraystretch}{1.70}
\small
\begin{tabular}{|p{1.6in}|p{0.5in}|p{0.3in} p{0.3in} p{0.45in}||p{0.5in}|p{0.3in} p{0.3in} p{0.45in}|}\hline
\multirow{2}{*}{\textbf{Topic}} &\multicolumn{4}{c}{\textbf{CRM Feedback}} &\multicolumn{4}{c|}{\textbf{Twitter}} \\\cline{2-9}
&Total\newline (approx) &Bus &Rail &Other / Generic &Total\newline (approx) &Bus &Rail &Other / Generic \\
\hline
Cleanliness / Maintenance &13,000 &24\% &70\% &6\% &34,000 &6\% &43\% &52\% \\
Crime / Harassment &4,000 &32\% &59\% &8\% &11,000 &7\% &26\% &66\% \\
Crowding &3,000 &62\% &38\% &0\% &6,000 &15\% &55\% &29\% \\
Customer Service &25,000 &44\% &42\% &14\% &105,000 &7\% &18\% &75\% \\
Fare Evasion &500 &17\% &82\% &1\% &2,000 &7\% &12\% &81\% \\
General &5,000 &35\% &34\% &31\% &31,000 &8\% &18\% &74\% \\
Information Provision &9,000 &29\% &65\% &6\% &34,000 &11\% &51\% &38\% \\
Route Suggestions &2,000 &88\% &10\% &3\% &1,000 &29\% &20\% &51\% \\
Operations &89,000 &83\% &12\% &5\% &102,000 &24\% &29\% &47\% \\
SmarTrip Cards / Fares &25,000 &18\% &24\% &58\% &12,000 &10\% &22\% &68\% \\
Unsafe Driving &6,000 &85\% &3\% &11\% &19,000 &4\% &93\% &3\% \\\hline
\textbf{Total (All Topics)} &181,500 &59\% &26\% &15\% &357,000 &12\% &32\% &56\% \\
\hline
\end{tabular}
\end{threeparttable}\end{adjustwidth}

\newpage
\subsubsection{Generating Performance Reporting Metrics from Customer Feedback}

\par
\begin{figure}[!h]
  \centering
  \begin{subfigure}{.49\textwidth}
    \includegraphics[width=\textwidth]{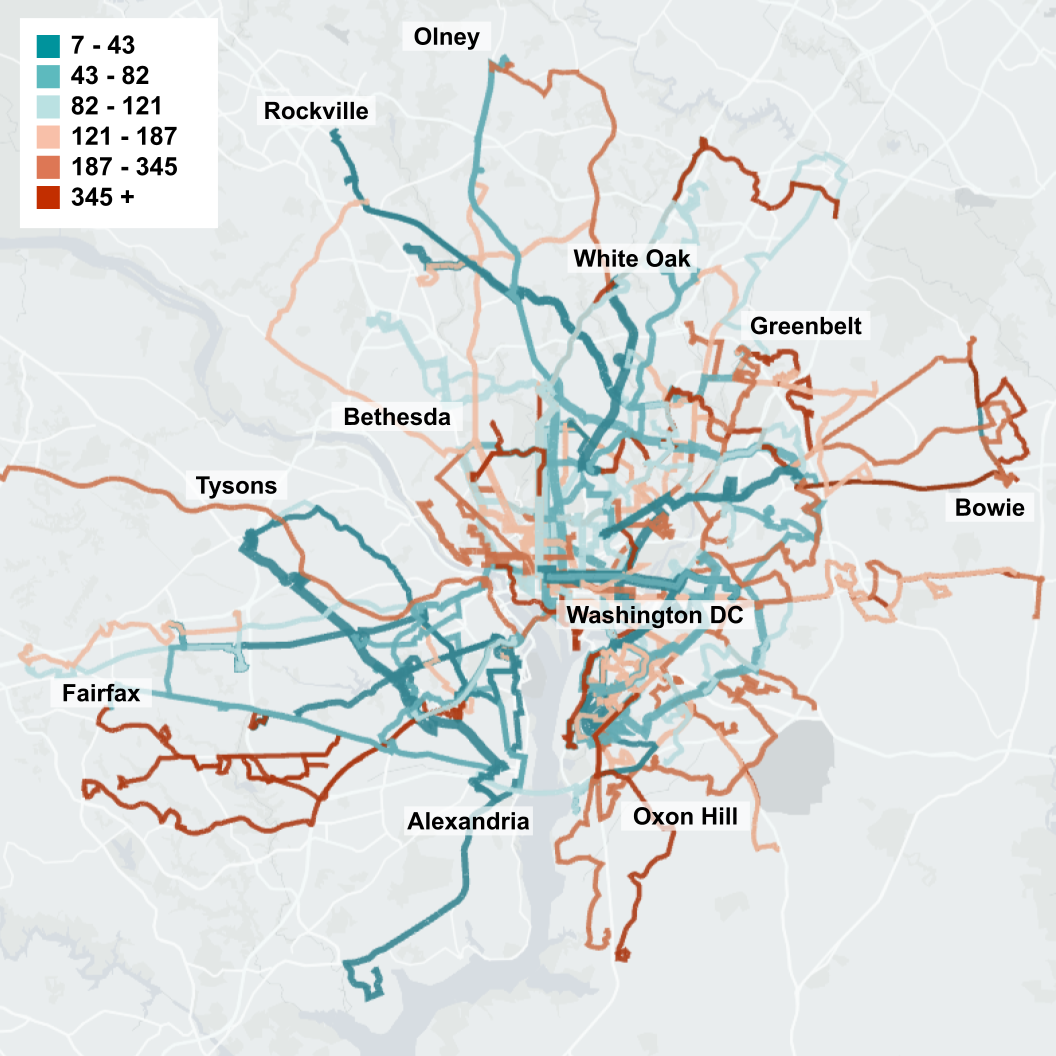}
    \caption{Operations-related by Metrobus routes.}
    \label{fig:buses}
  \end{subfigure}
  \begin{subfigure}{0.49\textwidth}
    \includegraphics[width=\textwidth]{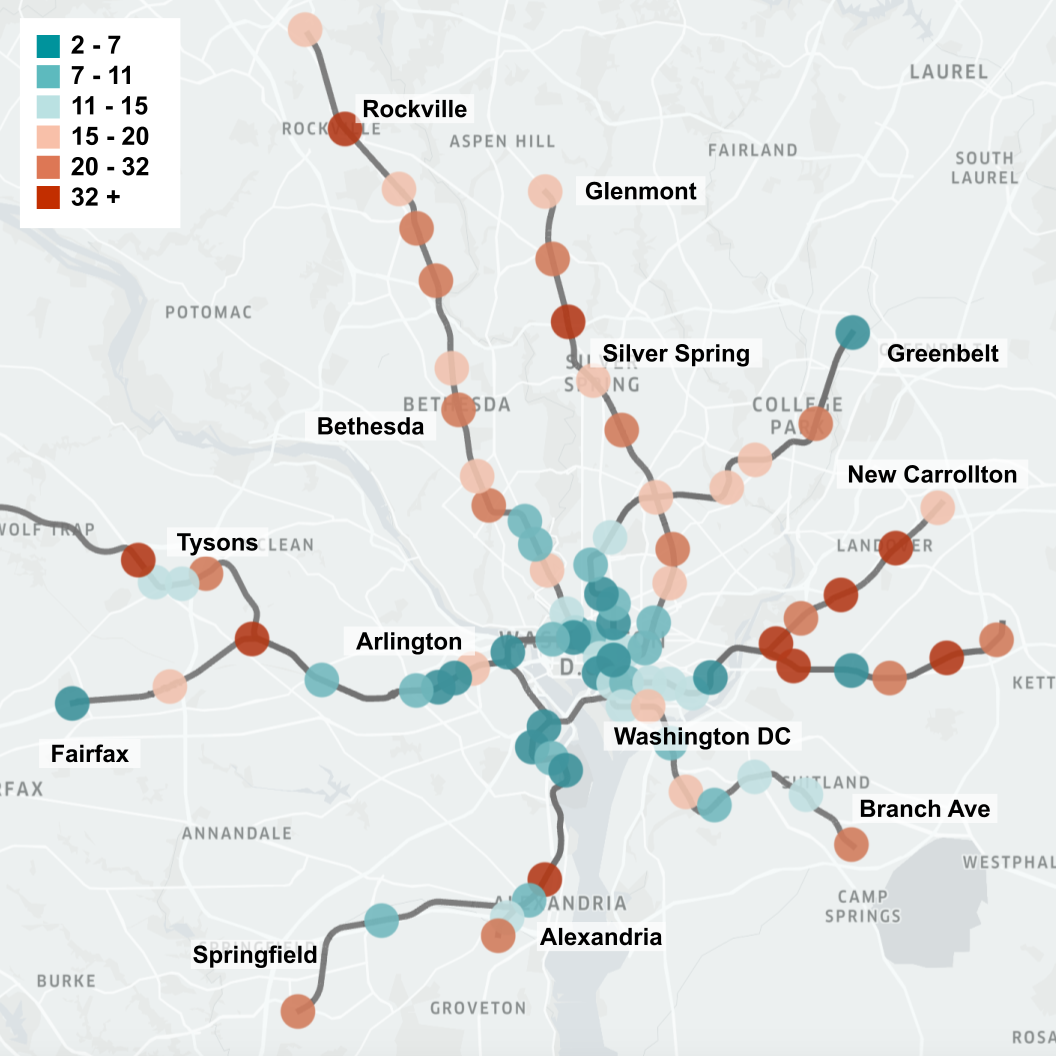}
    \caption{Cleanliness-related by Metrorail stations.}
    \label{fig:stations}
  \end{subfigure}
  \caption{{\label{fig:spatial} Spatial pattern of select topic complaints per million WMATA riders.}}
\end{figure}

 Lastly, we provide examples of how the structure added to open-ended text feedback using our model and proposed extended text analysis framework can be utilized to conduct cross-sectional, longitudinal, and strategic studies leveraging customer feedback as a performance indicator of the customer experience on various aspects of public transit systems. For the purpose of this demonstration, the reporting metric used is \emph{complaints per million riders}, where the volume of complaints is normalized by ridership data from WMATA's \emph{Trace} farecard database during the same analysis timeframe. This allows feedback levels to be compared across different time periods where system ridership varied, especially during the COVID-19 period. Figures \ref{fig:buses} and \ref{fig:stations} illustrate examples of cross-sectional studies showing the distribution of customer feedback on operations across WMATA's Metrobus routes and cleanliness in Metrorail stations over the year 2022, demonstrating the value-added of enabling topic-specific geospatial trends to be aggregated and visualized.


\bigskip
\begin{figure}[!h]
  \centering
  \begin{subfigure}{0.49\textwidth}
    \includegraphics[width=\textwidth]{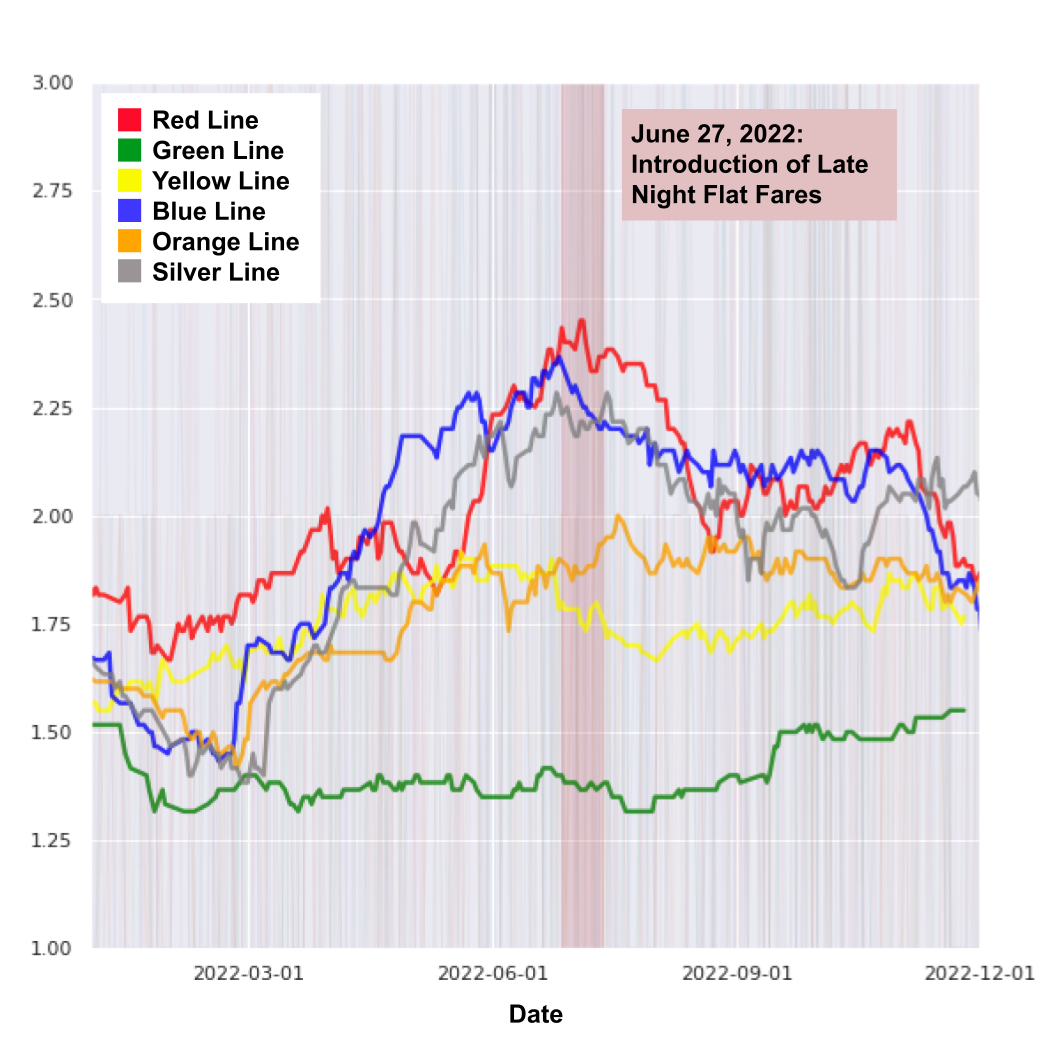}
    \caption{Fares-related by Metrorail route}
    \label{fig:fares}
  \end{subfigure}
  \hfill
  \begin{subfigure}{0.49\textwidth}
    \includegraphics[width=\textwidth]{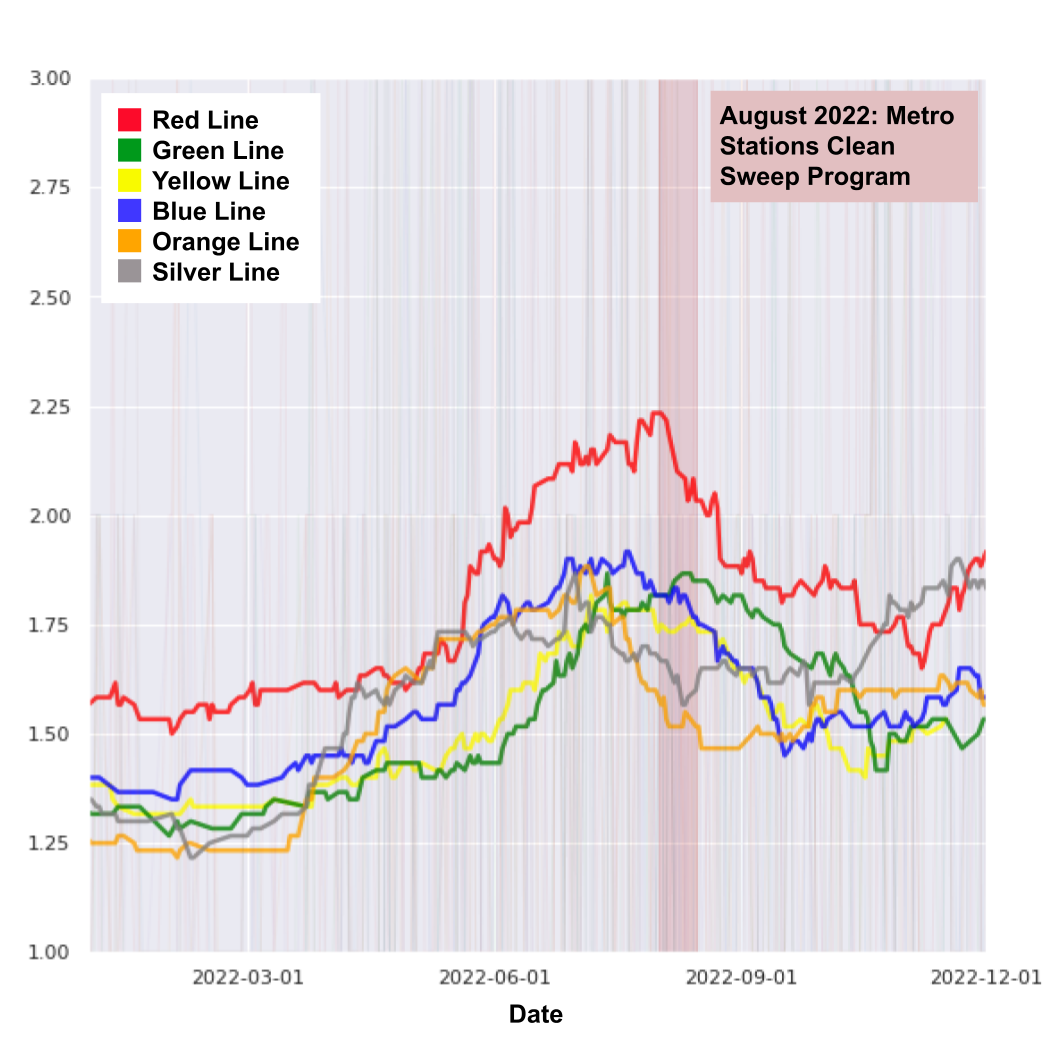}
    \caption{Cleanliness-related by Metrorail route.}
    \label{fig:cleanliness}
  \end{subfigure}
  \caption{{\label{fig:temporal} Temporal pattern of select topic complaints per million WMATA riders.}}
\end{figure}

Figures \ref{fig:fares} and \ref{fig:cleanliness} demonstrate examples of topic-specific longitudinal studies on complaints (illustrated as 30-day moving averages) about fares and cleanliness respectively during the year 2022. Using MetRoBERTa's transit topic classification, we are instantly able to observe that around the time WMATA implemented a late-night flat fare regime on Metrorail service starting June 27, 2022, fare-related complaints on most Metrorail lines decreased. Similarly, after a deep cleaning program for rail stations in August 2022, complaints about cleanliness on the rail system decreased on all lines.

These examples demonstrate what the authors envision as standard implementations of the MetRoBERTa extended analysis pipeline, where specific types of studies can be tailored to match the characteristics of the underlying data. Data with sparse aggregations, such as niche topics, feedback for specific bus routes, or smaller Metrorail stations, are more suitable for cross-sectional studies over long periods of time, where more opportunities for data collection may lead to meaningful comparison and analysis. Conversely, assets with rich aggregations which vary on a daily or weekly basis, such as complaints about specific popular rail lines, topics, or system-wide initiatives like fares and service changes are more suitable for longitudinal analysis (e.g. time series) where certain trends and anomalies can be identified.

\section{Conclusions}
In this study, we proposed and evaluated the value of leveraging customer feedback provided to transit agencies through traditional CRM systems to develop and deploy a transit-topic-aware large language model. The language model we developed as a part of this study, \emph{MetRoBERTa}, has proven capable of accurately classifying open-ended text feedback to relevant transit-specific topics. The model outperformed classical machine learning approaches that utilize lexicons and keyword-based representations, with an average transit topic classification accuracy of at least 90\%. We concluded this study by providing a value proposition of utilizing our language model, alongside additional text processing tools, to add structure to open-ended text sources of feedback like Twitter.

Based on our observations from this study, we expect \emph{MetRoBERTa's} performance to improve as it is retrained with more customer feedback data from WMATA or other sources. The literature indicates that the data-hungry nature of LLMs due to the massive number of parameters (123 million parameters for our RoBERTa-based model) does create a need for continued re-training as feasible to improve performance and ensure generalizability \cite{gururangan2020don}. Re-training our model from the current checkpoint can be completed in $\sim$30 minutes for every 10,000 feedback instances with a commercial-grade GPU (NVIDIA Tesla V100 or similar). A limitation of this study was the need to generate ground truth labels for the training dataset using LDA. While LDA or any other unsupervised or semi-supervised method can provide a reliable clustering of the topics, it can be observed in the examples provided in Table \ref{tab:misclassification} that the lexicon-based nature of clustering topics can lead to assigning customer comments into a false category, and such errors can cascade when that label is used to train a model. The vast majority of complaints, however, are categorized into LDA topics that are relevant to their manually assigned \emph{problem category} as shown in Appendix II and the subsequent model performance provides assurance that such misclassifications do not undermine model training in a substantial manner. Future work will investigate utilizing methods that account for label uncertainty \cite{northcutt2021confident}. Future work will also look into creating sub-models (smaller LLMs or lexicon-based models) to break down the broad categories detected by \emph{MetRoBERTa} into more granular topics which would create more value.

While the case study presented in this paper focused on comparing feedback data from traditional CRM to Twitter, the framework applies to all types of open-ended data that a transit agency might be interested in analyzing, including responses from ridership surveys, and incident and maintenance reports. In addition to the robust and accurate topic classification, the identification, to the extent possible, of transit agency assets mentioned in the customer feedback processed through our framework allows seamless integration with the agency's ridership and GTFS data, allowing for robust visualization and reporting of aggregate metrics and trends. Such a level of integration addresses a fundamental gap impeding the broader utilization of natural language processing in transit, creating a pathway for an automated, generalizable approach for ingesting, visualizing, and reporting transit riders' feedback at scale and empowering agencies to better understand and improve customer experience. The authors are working with WMATA on deploying this work for regular use at the agency and will continue to further develop the model backend and the use cases from processed data to best support the agency's customer experience analytics.




\section{Acknowledgements}
The authors would like to thank WMATA for their generous support of this project, and the MIT SuperCloud and Lincoln Laboratory Supercomputing Center for providing high-performance computing resources that have contributed to the research results reported in this paper.

\section{Author Contribution Statement}
The authors confirm their contribution to the paper as follows: study conception and design: A.A., M.L., D.P., G.P., M.E.; data collection: M.L., A.A., J.H.; language model development: A.A., J.H., M.L.; analysis and interpretation of results: M.L., A.A., D.P., G.P., A.H.; draft manuscript preparation: A.A., M.L., J.H., D.P., G.P., M.E., J.Z. All authors reviewed the results and approved the final version of the manuscript. The authors do not have any conflicts of interest to declare.


\nolinenumbers
\bibliographystyle{trb}
\bibliography{references}

\begin{thebibliography}{25}
\providecommand{\natexlab}[1]{#1}

\bibitem[{Weinstein(2022)}]{masstransit2022}
Weinstein, A., Diverse strategies in customer experience programs across North
  America help increase ridership, improve morale. \emph{Mass Transit
  Magazine}, 2022.

\bibitem[{Marolt et~al.(2015)Marolt, Pucihar, and
  Zimmermann}]{marolt2015social}
Marolt, M., A.~Pucihar, and H.-D. Zimmermann, Social CRM adoption and its
  impact on performance outcomes: A literature review. \emph{Organizacija},
  Vol.~48, No.~4, 2015, pp. 260--271.

\bibitem[{Nguyen et~al.(2015)Nguyen, Shirai, and Velcin}]{nguyen2015stocks}
Nguyen, T.~H., K.~Shirai, and J.~Velcin, Sentiment analysis on social media for
  stock movement prediction. \emph{Expert Systems with Applications}, Vol.~42,
  No.~24, 2015, pp. 9603--9611.

\bibitem[{Rane et~al.(2018)Rane, Kumar, and Kumar}]{Rane_2018airlines}
Rane, A., A.~Kumar, and A.~Kumar, Sentiment Classification System of Twitter
  Data for US Airline Service Analysis. \emph{2018 IEEE 42nd Annual Computer
  Software and Applications Conference (COMPSAC)}, 2018.

\bibitem[{Philander and Zhong(2016)}]{Philander_2016hospitality}
Philander, K.~S. and Y.~Y. Zhong, Twitter sentiment analysis: Capturing
  sentiment from integrated resort tweets. \emph{International Journal of
  Hospitality Management}, 2016.

\bibitem[{Watkins et~al.(2021)Watkins, Berrebi, Erhardt, Hoque, Goyal,
  Brakewood, Ziedan, Darling, Hemily, and Kressner}]{watkins2021recent}
Watkins, K., S.~Berrebi, G.~Erhardt, J.~Hoque, V.~Goyal, C.~Brakewood,
  A.~Ziedan, W.~Darling, B.~Hemily, and J.~Kressner, Recent Decline in Public
  Transportation Ridership: Analysis, Causes, and Responses. \emph{TCRP
  Research Report}, , No. 231, 2021.

\bibitem[{O’Brien(2023)}]{aptwitter}
O’Brien, M., Elon Musk put new limits on tweets. Users and advertisers might
  go elsewhere. \emph{Associated Press}, 2023.

\bibitem[{Lock and Pettit(2020)}]{Lock_2020}
Lock, O. and C.~Pettit, Social media as passive geo-participation in
  transportation planning – how effective are topic modeling \& sentiment
  analysis in comparison with citizen surveys? \emph{Geo-spatial Information
  Science}, 2020.

\bibitem[{Das and Zubaidi(2021)}]{Das_2021}
Das, S. and H.~A. Zubaidi, City Transit Rider Tweets: Understanding Sentiments
  and Politeness. \emph{Journal of Urban Technology}, 2021.

\bibitem[{Hosseini et~al.(2018)Hosseini, El-Diraby, and
  Shalaby}]{Hosseini_2018}
Hosseini, M., T.~E. El-Diraby, and A.~Shalaby, Supporting sustainable system
  adoption: Socio-semantic analysis of transit rider debates on social media.
  \emph{Sustainable Cities and Society}, 2018.

\bibitem[{Kabbani et~al.(2022)Kabbani, Klumpenhouwer, El-Diraby, and
  Shalaby}]{Kabbani_2022}
Kabbani, O., W.~Klumpenhouwer, T.~E. El-Diraby, and A.~Shalaby, What do riders
  say and where? The detection and analysis of eyewitness transit tweets.
  \emph{Journal of Intelligent Transportation Systems}, 2022.

\bibitem[{Haghighi et~al.(2018)Haghighi, Liu, Wei, Wei, Wei, Wei, Li, and
  Shao}]{Haghighi_2018}
Haghighi, N., X.~C. Liu, R.~Wei, R.~Wei, R.~Wei, R.~Wei, W.~Li, and H.~Shao,
  Using Twitter data for transit performance assessment: a framework for
  evaluating transit riders’ opinions about quality of service. \emph{Public
  Transport}, 2018.

\bibitem[{Al-Sahar et~al.(2023)Al-Sahar, Klumpenhouwer, Shalaby, and
  El-Diraby}]{al2023using}
Al-Sahar, R., W.~Klumpenhouwer, A.~Shalaby, and T.~El-Diraby, Using Twitter to
  Gauge Customer Satisfaction Response to a Major Transit Service Change in
  Calgary, Canada. \emph{Transportation Research Record}, 2023, p.
  03611981231179167.

\bibitem[{Liu(2019)}]{Liu_2019}
Liu, X.~C., Webinar: Social Transportation Analytic Toolbox (STAT) for Transit
  Networks, 2019.

\bibitem[{Mendez et~al.(2019)Mendez, Lobel, Parra, and
  Herrera}]{Mendez_2019chile}
Mendez, J.~T., H.~Lobel, D.~Parra, and J.~C.~B. Herrera, Using Twitter to Infer
  User Satisfaction With Public Transport: The Case of Santiago, Chile.
  \emph{IEEE Access}, 2019.

\bibitem[{Luong and Houston(2015)}]{Luong_2015la}
Luong, T.~T. and D.~Houston, Public opinions of light rail service in Los
  Angeles, an analysis using Twitter data. \emph{null}, 2015.

\bibitem[{Arjona et~al.(2020)Arjona, Osorio-Arjona, Horak, Svoboda, and
  García-Ruiz}]{Arjona_2020madrid}
Arjona, J.~O., J.~Osorio-Arjona, J.~Horak, R.~Svoboda, and Y.~García-Ruiz,
  Social media semantic perceptions on Madrid Metro system: Using Twitter data
  to link complaints to space. \emph{Sustainable Cities and Society}, 2020.

\bibitem[{Devlin et~al.(2018)Devlin, Chang, Lee, and
  Toutanova}]{devlin2018bert}
Devlin, J., M.-W. Chang, K.~Lee, and K.~Toutanova, Bert: Pre-training of deep
  bidirectional transformers for language understanding. \emph{arXiv preprint
  arXiv:1810.04805}, 2018.

\bibitem[{Radford et~al.(2019)Radford, Wu, Child, Luan, Amodei, Sutskever
  et~al.}]{radford2019gpt}
Radford, A., J.~Wu, R.~Child, D.~Luan, D.~Amodei, I.~Sutskever, et~al.,
  Language models are unsupervised multitask learners. \emph{OpenAI blog},
  Vol.~1, No.~8, 2019, p.~9.

\bibitem[{Liu et~al.(2019)Liu, Ott, Goyal, Du, Joshi, Chen, Levy, Lewis,
  Zettlemoyer, and Stoyanov}]{liu2019roberta}
Liu, Y., M.~Ott, N.~Goyal, J.~Du, M.~Joshi, D.~Chen, O.~Levy, M.~Lewis,
  L.~Zettlemoyer, and V.~Stoyanov, Roberta: A robustly optimized bert
  pretraining approach. \emph{arXiv preprint arXiv:1907.11692}, 2019.

\bibitem[{Antypas et~al.(2022)Antypas, Ushio, Camacho-Collados, Silva, Neves,
  and Barbieri}]{antypas2022twitter}
Antypas, D., A.~Ushio, J.~Camacho-Collados, V.~Silva, L.~Neves, and
  F.~Barbieri, {T}witter Topic Classification. In \emph{Proceedings of the 29th
  International Conference on Computational Linguistics}, International
  Committee on Computational Linguistics, Gyeongju, Republic of Korea, 2022,
  pp. 3386--3400.

\bibitem[{Barbieri et~al.(2020)Barbieri, Camacho-Collados, Espinosa~Anke, and
  Neves}]{barbieri-etal-2020-tweeteval}
Barbieri, F., J.~Camacho-Collados, L.~Espinosa~Anke, and L.~Neves,
  {T}weet{E}val: Unified Benchmark and Comparative Evaluation for Tweet
  Classification. In \emph{Findings of the Association for Computational
  Linguistics: EMNLP 2020}, Association for Computational Linguistics, Online,
  2020, pp. 1644--1650.

\bibitem[{Shuman et~al.(2022)Shuman, Abdelhalim, Stewart, Campbell, Patel,
  S{\'a}nchez~de Madariaga, and Zhao}]{shuman2022inferring}
Shuman, D., A.~Abdelhalim, A.~F. Stewart, K.~B. Campbell, M.~Patel,
  I.~S{\'a}nchez~de Madariaga, and J.~Zhao, Inferring Mobility of Care Travel
  Behavior From Transit Origin-Destination Data. \emph{arXiv preprint
  arXiv:2211.04915}, 2022.

\bibitem[{Gururangan et~al.(2020)Gururangan, Marasovi{\'c}, Swayamdipta, Lo,
  Beltagy, Downey, and Smith}]{gururangan2020don}
Gururangan, S., A.~Marasovi{\'c}, S.~Swayamdipta, K.~Lo, I.~Beltagy, D.~Downey,
  and N.~A. Smith, Don't stop pretraining: Adapt language models to domains and
  tasks. \emph{arXiv preprint arXiv:2004.10964}, 2020.

\bibitem[{Northcutt et~al.(2021)Northcutt, Jiang, and
  Chuang}]{northcutt2021confident}
Northcutt, C., L.~Jiang, and I.~Chuang, Confident learning: Estimating
  uncertainty in dataset labels. \emph{Journal of Artificial Intelligence
  Research}, Vol.~70, 2021, pp. 1373--1411.

\end{thebibliography}
\newpage
\section{Appendix I - List of WMATA CRM Problem Categories}
\vspace{50pt}
\begin{multicols}{2}
\begin{itemize}
    \item Accident
    \item Accessibility
    \item Accidental Door Closing
    \item Advertising
    \item Announcements
    \item Bicycles
    \item Bus Info Display
    \item Cleaning
    \item Cooling
    \item COVID-19
    \item Commendation
    \item Crime
    \item Crowding
    \item Curbing Bus
    \item Delays
    \item Did Not Follow Route
    \item Did Not Stop
    \item Door Closing
    \item Early
    \item Fare Evasion
    \item Fare Increase
    \item Fare Info Display
    \item Fare Value Missing
    \item Farebox
    \item Following Procedures
    \item General Information
    \item Go Card
    \item Harassment
    \item Heat
    \item Incorrect Fare
    \item Incorrect Parking Fee
    \item Inaccurate Information
    \item Injury
    \item Lighting
    \item Literature
    \item Maintenance
    \item MetroAccess
    \item Mobile Feedback
    \item No Show
    \item Other
    \item Parking Permits
    \item Passenger Assistance
    \item Platform Info Display
    \item Property Damage
    \item Request Additional Service
    \item Restrooms
    \item Rude Employee
    \item Rush Hour Promise
    \item Safety
    \item Seating
    \item Security
    \item Service Changes
    \item Service Disruption
    \item SmarTrip Card
    \item Station Entry Info Display
    \item Suggestion
    \item Tickets/Fines
    \item Train Info Display
    \item Trip Planner
    \item Unprofessional Behavior
    \item Unprofessional Tone
\end{itemize}
\end{multicols}

\newpage
\section{Appendix II - Ratios of WMATA CRM Problem Categories in LDA-detected Topics}

\begin{table}[!h]
  \centering
  \renewcommand{\arraystretch}{1.35}
    \begin{tabular}{|l|c|c|}
    \hline
    \textbf{LDA Topic} & \textbf{CRM Problem Category} & \textbf{\% of Topic} \\
    \hline
    \multirow{3}[6]{*}{\textbf{Cleanliness / Maintenance}} & \textit{Maintenance} & 0.29 \\
\cline{2-3}          & \textit{Cleaning} & 0.17 \\
\cline{2-3}          & \textit{Cooling} & 0.11 \\
    \hline
    \multirow{3}[6]{*}{\textbf{Crime / Harassment / Security}} & \textit{Crime} & 0.52 \\
\cline{2-3}          & \textit{Harassment} & 0.33 \\
\cline{2-3}          & \textit{Security} & 0.15 \\
    \hline
    \textbf{Crowding} & \textit{Crowding} & 1.00 \\
    \hline
    \multirow{3}[6]{*}{\textbf{Customer Service}} & \textit{Commendation} & 0.42 \\
\cline{2-3}          & \textit{Rude Employee} & 0.16 \\
\cline{2-3}          & \textit{Unprofessional Behavior} & 0.06 \\
    \hline
    \textbf{Fare Evasion} & \textit{Fare Evasion} & 1.00 \\
    \hline
    \multirow{3}[6]{*}{\textbf{General}} & \textit{General Information} & 0.14 \\
\cline{2-3}          & \textit{Other} & 0.10 \\
\cline{2-3}          & \textit{Commendation} & 0.08 \\
    \hline
    \multirow{3}[6]{*}{\textbf{Information Provision}} & \textit{Delays} & 0.31 \\
\cline{2-3}          & \textit{Service Disruption} & 0.08 \\
\cline{2-3}          & \textit{General Information} & 0.07 \\
    \hline
    \multirow{3}[6]{*}{\textbf{Operation / Delay / Procedures}} & \textit{Did Not Stop} & 0.22 \\
\cline{2-3}          & \textit{No Show} & 0.19 \\
\cline{2-3}          & \textit{Delays} & 0.17 \\
    \hline
    \multirow{3}[6]{*}{\textbf{Route Suggestions}} & \textit{Bus Info Display} & 0.66 \\
\cline{2-3}          & \textit{Delays} & 0.07 \\
\cline{2-3}          & \textit{Suggestion} & 0.04 \\
    \hline
    \multirow{3}[6]{*}{\textbf{SmarTrip Card/Fares}} & \textit{SmarTrip Card} & 0.46 \\
\cline{2-3}          & \textit{Fare Value Missing} & 0.20 \\
\cline{2-3}          & \textit{Other} & 0.05 \\
    \hline
    \multirow{3}[6]{*}{\textbf{Unsafe Driving / Vehicle Conflict on Road}} & \textit{Unsafe Operation} & 0.69 \\
\cline{2-3}          & \textit{Safety} & 0.12 \\
\cline{2-3}          & \textit{Rude Employee} & 0.03 \\
    \hline
    \end{tabular}%
  \label{tab:addlabel}%
\end{table}%

\end{document}